\documentclass[letterpaper, 10 pt, conference]{ieeeconf}  
\usepackage{url}
\usepackage{xcolor}

\IEEEoverridecommandlockouts                              

\usepackage{graphics} 
\usepackage{epsfig} 
\usepackage{mathptmx} 
\usepackage{times} 
\usepackage{amsmath} 
\usepackage{amssymb}  
\usepackage{booktabs}
\usepackage{multirow}
\usepackage{graphicx}
\usepackage{tikz}
\usetikzlibrary{arrows.meta,positioning,fit,backgrounds,calc}
\usepackage{xcolor}
\usepackage{hyperref}
\usepackage[caption=false,font=footnotesize]{subfig}

\definecolor{cInput}{RGB}{243,247,252}
\definecolor{cCore}{RGB}{232,241,250}
\definecolor{cCoreDark}{RGB}{210,228,244}
\definecolor{cCtrl}{RGB}{241,235,252}
\definecolor{cCtrlDark}{RGB}{223,212,246}
\definecolor{cAction}{RGB}{252,244,232}
\definecolor{cLoop}{RGB}{245,245,245}
\definecolor{cAccent}{RGB}{70,110,170}

\title{\LARGE \bf
IMPACT-HOI: Supervisory Control for Onset-Anchored Partial HOI Event Construction
}

\author{Haoshen Zhang$^{1\dagger}$, Di Wen$^{1\dagger*}$, Kunyu Peng$^{1,2}$, David Schneider$^{1}$, Zeyun Zhong$^{1}$, Alexander Jaus$^{1}$,\\ Zdravko Marinov$^{1}$,Jiale Wei$^{1}$, Ruiping Liu$^{1}$, Junwei Zheng$^{1,3}$, Yufan Chen$^{1}$, Yufeng Zhang $^{1}$, \\ Yuanhao Luo$^{1}$, Lei Qi$^{4}$,and Rainer Stiefelhagen$^{1}$%
\thanks{$^{\dagger}$These authors contributed equally to this work.}%
\thanks{$^{*}$Corresponding author: Di Wen ({\tt\small di.wen@kit.edu}).}%
\thanks{$^{1}$The authors are with Karlsruhe Institute of Technology, 76131 Karlsruhe, Germany.}%
\thanks{$^{2}$The author is with INSAIT, Sofia University “St. Kliment Ohridski”, 1784 Sofia, Bulgaria.}%
\thanks{$^{3}$The author is with ETH Zurich, 8092 Zurich, Switzerland.}%
\thanks{$^{4}$The author is with Technical University of Munich, 80333 Munich, Germany.}%
}

\begin{document}

\maketitle
\thispagestyle{empty}
\pagestyle{empty}

\begin{abstract}
We present IMPACT-HOI, a mixed-initiative framework for
annotating egocentric procedural video by constructing
structured event graphs for Human-Object Interactions (HOI),
motivated by the need for high-quality structured supervision
for learning robot manipulation from human demonstration.
IMPACT-HOI frames this task as the incremental resolution of
a partially specified, onset-anchored event state.
A trust-calibrated controller selects among direct queries,
human-confirmed suggestions, and conservative completions
based on empirical annotator behavior and evidence quality.
A risk-bounded execution protocol, utilizing atomic rollback,
ensures that human-confirmed decisions are preserved against
conflicting automated updates.
A user study with 9 participants shows a 13.5\% reduction in
manual annotation actions, a 46.67\% event match rate, and
zero confirmed-field violations under the studied protocol.
The code will be made publicly available at
\href{https://github.com/541741106/IMPACT\_HOI}%
{https://github.com/541741106/IMPACT\_HOI}.
\end{abstract}
\section{INTRODUCTION}

The promise of learning robot manipulation from human
demonstration relies on the premise that if we can record
human-object interactions, agents can learn to replicate
them. While recent advancements in imitation learning and
contact-rich policy learning suggest this is
achievable~\cite{chi2025visuomotor}, producing high-quality
supervision at scale remains a challenge. A video alone is
insufficient; what is needed is a verified, structured record
specifying which hand performed which action on which object
and when functional contact occurred. This temporal grounding
is crucial for conditioning contact-aware reward signals,
state-change supervision, and bimanual coordination
priors~\cite{kareer2025egomimic}. While egocentric
procedural video, such as that from
Ego4D~\cite{grauman2022ego4d} and
EPIC-KITCHENS~\cite{darkhalil2022epic}, contains this
information, the bottleneck lies in the structured annotation
required to make it actionable.

Constructing HOI event records is challenging due to
interdependent fields (temporal boundaries, semantic verb,
instrument role, and target role) whose resolution costs and
verification risks vary. Incorrectly confirming one field
can propagate errors across events, and in dense, bimanual
interactions, sequential annotation becomes inefficient and
fragile.
Existing methods are inadequate: end-to-end HOI
detectors~\cite{kim2021hotr,tamura2021qpic,liao2022gen}
assume complete input and output with no partial event state
or human oversight, while platforms like
CVAT~\cite{cvat2023} and VIA~\cite{dutta2019via} do not
prioritize unresolved fields or protect confirmed decisions
from modification. The central challenge is therefore not
HOI recognition alone, but controllable human--machine
co-construction of structured event records under partial
observability and asymmetric authority.

We propose IMPACT-HOI, a mixed-initiative supervisory
framework for structured HOI annotation. IMPACT-HOI frames
annotation as Lock-aware Partial Event Completion (LPEC):
it operates on a partially specified per-hand event state,
resolving only open variables while preserving
human-confirmed decisions through lock-constrained decoding.
The system's temporal anchor, the Hand-guided Onset Prior
(HOP), uses hand-motion evidence to localize
functional-contact onset and provide temporal context for
completion. A lightweight event-local completion module
jointly predicts onset and semantics, followed by
Statistics-guided Cooperative Refinement (SCR) to ensure
cross-field consistency using empirical priors and
interaction history. Finally, a Trust-Calibrated Supervisory
Controller (TSC) ranks candidate interventions and assigns
machine authority based on empirically calibrated acceptance
and cost signals. Safe mixed-initiative operation is ensured
through lock-constrained decoding, authority-bounded
execution, and structured logging for recalibration.

The contributions of this work are as follows.
\begin{itemize}
\item \textbf{Onset-anchored event-local completion
  pipeline.} We introduce \emph{Hand-guided Onset Prior
  (HOP)} and a shared completion module to resolve onset
  and event semantics from partial states and visual cues.

\item \textbf{Statistics-guided consistency refinement.} We
  propose \emph{Statistics-guided Cooperative Refinement
  (SCR)} to improve consistency using priors, co-occurrence
  statistics, and verb-specific tendencies.

\item \textbf{Trust-calibrated supervisory controller for
  mixed-initiative annotation.} We develop
  \emph{Trust-Calibrated Supervisory Controller (TSC)} to
  rank interventions and assign machine authority,
  demonstrating a 13.5\% reduction in manual actions and
  46.67\% event match rate with no confirmed-field
  violations under the studied protocol.
\end{itemize}
\section{RELATED WORK}
\label{sec:related}

\noindent\textbf{Video Human--Object Interaction and Per-Hand Interaction Understanding}
\label{sec:rw_videohoi}
Video HOI has evolved from clip-level relational models~\cite{sunkesula2020lighten, wang2021stigpn, ji2021videohor} to transformer detectors with tubelet tokens~\cite{tu2022tutor}; ST-HOI introduced VidHOI and diagnosed feature-trajectory misalignment~\cite{chiou2021sthoi}. ASSIGN models asynchronous, sparse entity lifecycles~\cite{morais2021assign}, supporting our per-hand view; HOI-DA unifies detection and anticipation via residual pair-state transitions~\cite{luo2026rethinking}; RoHOI provides an image-level corruption benchmark for robustness evaluation~\cite{wen2025rohoi}.
Large-scale benchmarks~\cite{zhu2026egosound, zhang2025egonight, damen2022epic} offer fine-grained annotations: EPIC-KITCHENS-100 provides verb-noun factorization with narration~\cite{damen2022epic}; HOI4D offers per-hand and object pose~\cite{liu2022hoi4d}; ARCTIC contributes bimanual contact annotations~\cite{fan2023arctic}; VISOR provides hand-object masks~\cite{darkhalil2022epic}; Ego4D introduces state-change framing~\cite{grauman2022ego4d}. IMPACT targets industrial assembly with multi-view RGB-D, bimanual verb-noun labels, procedural hierarchies, anomaly taxonomy, and synchronized null spans~\cite{wen2026impact}, supporting downstream robot imitation~\cite{kareer2025egomimic} and visuomotor diffusion~\cite{chi2025visuomotor}. While prior work focuses on video \emph{inference}, we address incremental \emph{construction} of a per-hand event record from partial states under human locks.

\noindent\textbf{Temporal Structure, Onset Reasoning, and Fine-Grained Action Understanding}
\label{sec:rw_temporal}
Fine-grained action understanding is increasingly target-conditioned, such as referring atomic action recognition~\cite{peng2024ravar} and diffusion-based segmentation in multi-person video~\cite{peng2025hopadiff}. Anticipation work~\cite{zhong2023afft, ni2023hoigaze} uses onset-like cues for forecasting. Manousaki et al.~\cite{manousaki2024osca} predict post-states from observations up to an action's point-of-no-return, aligned with Ego4D's PNR framing~\cite{grauman2022ego4d}.
We condition on hand identity and treat functional-contact onset as a constraint for verb, noun, and boundary assignments. Open-set action recognition~\cite{peng2024openset} and noisy-label learning~\cite{xu2024noisesar} support modeling annotator input as imperfect.

\noindent\textbf{Human-in-the-Loop Supervisory Control and Industrial Procedural Assistants}
\label{sec:rw_hitl}
Closest to our authority-level assignment is work selecting the \emph{mode} of supervision: EVA-VOS picks per-frame annotation types via a PPO agent trading quality against cost~\cite{delatolas2024evavos}; we adopt the same utility-over-cost view but operate per event field with three authority levels (\texttt{human\_only}, \texttt{human\_confirm}, \texttt{safe\_local}).
Active learning for video~\cite{rana2023claus,singh2024ssalvad} and point-supervised temporal localisation~\cite{liu2025pstal} select samples or queries rather than intervention modes, and usage-time adaptation via LoRA updates~\cite{yang2026lit} or SAM-decoder tuning~\cite{schon2024sam} from user feedback motivates our parameter-efficient adapter; we, however, select interventions proactively rather than reactively.
Industrial assistants~\cite{wen2025mica,wen2025snap}, Assembly101~\cite{sener2022assembly101}, and IMPACT~\cite{wen2026impact} define a regime of on-device, bimanual, error-prone procedural supervision, within which we cast HOI construction as mixed-initiative control over a partial event state with field-level authority and risk-bounded completion.

\section{Methodology}
\label{sec:method}

\begin{figure*}[t]
\centering
\includegraphics[width=0.8\textwidth]{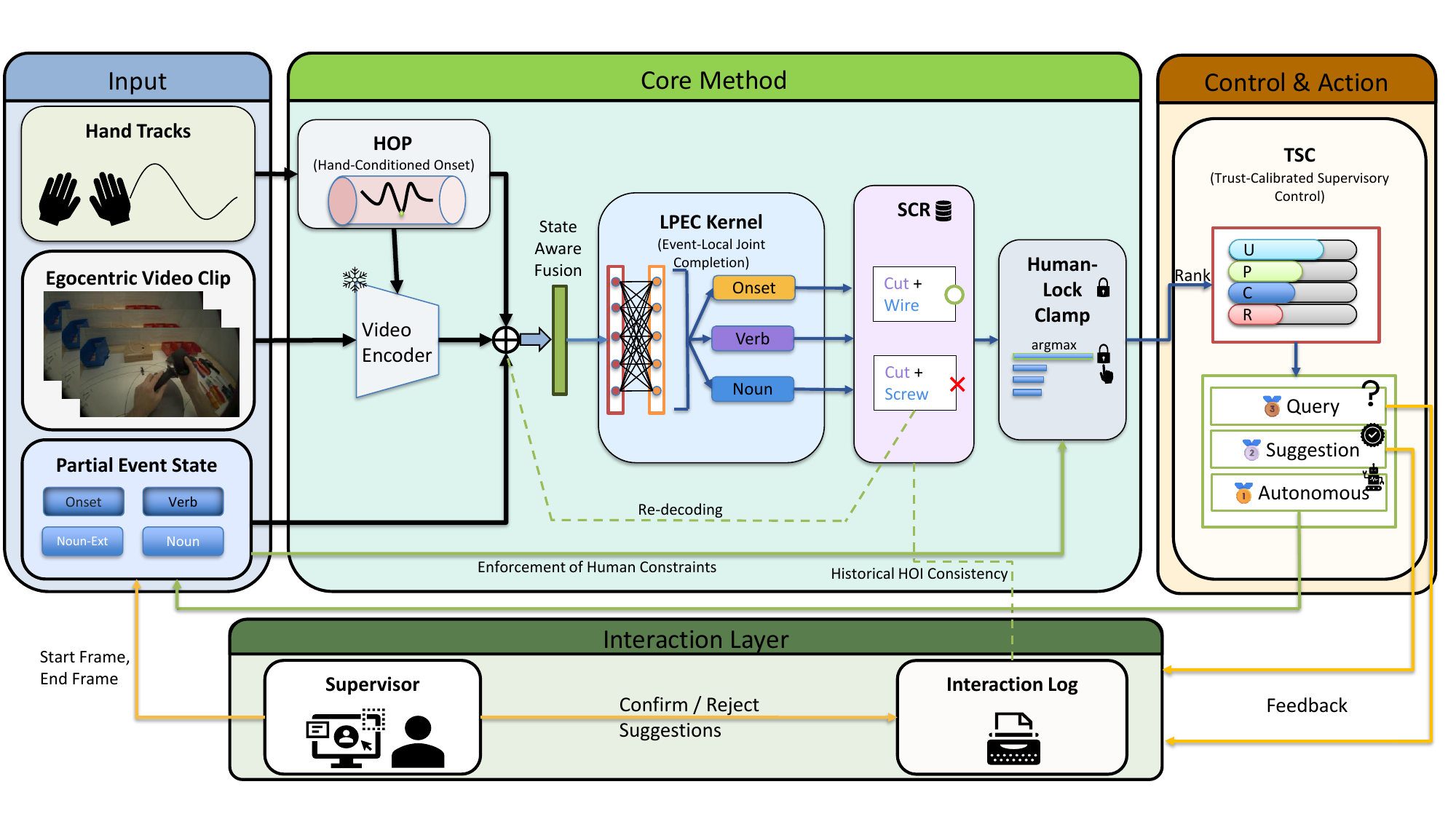}
\vskip-2ex
\caption{%
IMPACT-HOI starts with a partial event state and video evidence.
The Lock-aware Partial Event Completion (LPEC) module resolves
open fields through joint onset and semantic prediction, refined
by Statistics-guided Cooperative Refinement and re-decoding under
human-confirmed locks. The Trust-Calibrated Supervisory Controller
(TSC) selects actions, either eliciting human responses or applying
safe updates. Human responses and logged traces feed back into the
event state and controller, forming a closed annotation loop.}
\vskip-4ex
\label{fig:method_overview}
\end{figure*}

HOI annotation is not a recognition problem. Instead of predicting
unconstrained labels, the system works with an evolving partial
event state, resolving only open variables while preserving prior
human decisions. We frame HOI assistance as \emph{Lock-aware
Partial Event Completion (LPEC)} of partially specified per-hand
events under human locks.

\subsection{Problem Formulation}
\label{sec:formulation}

Let $\mathcal{H}=\{\mathrm{Left},\mathrm{Right}\}$ denote the two
hand streams, treated symmetrically. For a selected stream
$h\in\mathcal{H}$, the active event is
\begin{equation}
e_h=(h,\,t_s,\,t_o,\,t_e,\,v,\,n),
\label{eq:event}
\end{equation}
where $t_s$, $t_o$, and $t_e$ denote interaction start,
functional-contact onset, and interaction end, respectively, while
$v$ and $n$ denote the verb and noun. The editable runtime state is
\begin{equation}
x_h=(\tilde{e}_h,\,\sigma_h,\,\rho_h),
\label{eq:state}
\end{equation}
where $\tilde{e}_h$ stores the current field values, $\sigma_h$
stores per-field status (empty, suggested, or confirmed), and
$\rho_h$ stores provenance and lock metadata. Because $t_s$ and
$t_e$ are directly editable on the timeline, the event-local
completion target reduces to
\begin{equation}
y_h=(t_o,\,v,\,b,\,n),\qquad b\in\{0,1\},
\label{eq:target}
\end{equation}
where $b$ is the noun-existence indicator. The system resolves only
the still-open components of $y_h$, while preserving all confirmed
decisions and enforcing verb--noun validity. This partial-state
constrained completion view defines the core LPEC problem.

\subsection{Hand-Conditioned Onset Grounding}
\label{sec:grounding}

Semantic interpretation depends critically on where functional
contact begins. To reduce onset ambiguity, we introduce a
\emph{phase-aware Hand-guided Onset Prior (HOP)} that uses the
selected hand track to produce a structured temporal prior over the
current event interval rather than a single heuristic motion peak.

Let $W_h=[t_s,t_e]$ denote the current event interval for hand
stream $h$. When persistent hand tracks are available, each frame
$t\in W_h$ provides a hand state
\begin{equation}
\phi_t=(b_t,\,c_t,\,a_t,\,m_t,\,\eta_t),
\label{eq:handstate}
\end{equation}
where $b_t$ is the hand box, $c_t\in\mathbb{R}^2$ is its center,
$a_t$ is its area, $m_t$ is a smoothed hand-motion response, and
$\eta_t$ denotes handedness evidence. These framewise cues define
the hand-conditioned temporal evidence used by HOP.

Rather than selecting onset directly from motion alone, HOP first
derives a coarse semantic prior from an \emph{unguided} event-local
video summary computed without any onset band. This yields
\begin{equation}
\psi_h=(\hat{v}_h,\,g_h,\,\tau_h,\,\gamma_h),
\label{eq:semprior}
\end{equation}
where $\hat{v}_h$ is the highest-scoring coarse verb hypothesis,
$g_h\in\{\mathrm{boundary},\mathrm{early},\mathrm{mid},
\mathrm{late}\}$ is its phase family, $\tau_h\in[0,1]$ is the
corresponding family template ratio within the segment, and
$\gamma_h\in[0,1]$ is the family confidence. The template ratio
defines a target location
\begin{equation}
t_h^\star=t_s+\tau_h(t_e-t_s),
\label{eq:target_loc}
\end{equation}
around which HOP searches for phase-consistent onset candidates.

We construct a candidate set
\begin{equation}
\mathcal{C}_h=
\mathcal{C}_h^{\mathrm{boundary}}
\cup\mathcal{C}_h^{\mathrm{peak}}
\cup\mathcal{C}_h^{\mathrm{valley}}
\cup\mathcal{C}_h^{\mathrm{stab}},
\label{eq:candidates}
\end{equation}
whose elements correspond to boundary-aligned starts, local motion
peaks, low-motion contact valleys, and the starts of short
stabilization runs, respectively. Each candidate $c\in\mathcal{C}_h$
is scored by
\begin{equation}
s_h(c)=\Phi\!\left(c;\,g_h,\,t_h^\star,\,
\{m_t\}_{t\in W_h}\right),
\label{eq:score_hop}
\end{equation}
where $\Phi(\cdot)$ combines phase compatibility with $g_h$,
proximity to the template location $t_h^\star$, local motion
evidence, and short-range temporal support before and after the
candidate. This design allows HOP to favor different onset
mechanisms for different interaction families, e.g.,
segment-start boundaries for hold-like events, contact valleys for
contact-driven manipulations, and later stabilization patterns for
release or put-down events. Let $c_h^\star$ be the selected
candidate:
\begin{equation}
c_h^\star=\arg\max_{c\in\mathcal{C}_h}s_h(c).
\label{eq:best_candidate}
\end{equation}
Its reliability is summarized as
\begin{equation}
\kappa_h=\Omega\!\left(\nu_h,\,s_h(c_h^\star),\,
\delta_h,\,\rho_h\right)\in[0,1],
\label{eq:reliability}
\end{equation}
where $\nu_h$ aggregates track coverage, motion dominance, coarse
semantic confidence, and handedness purity; $\delta_h$ is the
margin over competing phase candidates; and $\rho_h$ measures
local frame support around $c_h^\star$. If coverage, handedness
purity, or calibrated confidence is insufficient, HOP abstains and
no hand prior is emitted.

The resulting prior is
\begin{equation}
\pi_h=(\hat{t}_o,\,B_h,\,\kappa_h),
\label{eq:prior}
\end{equation}
where $\hat{t}_o$ is the selected onset frame and
$B_h=[\ell_h,r_h]\subseteq W_h$ is a frame-level onset band
induced by the selected candidate, its supporting frame cluster,
and the semantic phase family. This prior does not overwrite the
editable event state. Instead, it conditions onset-local temporal
sampling of the frozen video backbone and the local onset window
used by downstream completion. When persistent hand evidence is
unavailable, or when HOP abstains, the system falls back to
timeline-derived temporal guidance alone.

\subsection{Event-Local Onset-Semantics Completion}
\label{sec:completion}

The completion target $y_h$ depends on the selected hand stream,
the current partial state, and the subset of fields that remain
open, making this fundamentally different from clip-level
recognition. We therefore resolve onset and semantics through a
shared event-local completion pipeline within LPEC.

\paragraph{Feature representation.}
Event-local inputs are grouped as
\begin{equation}
r_h=[s_h,\;q_h,\;u_h,\;z_h],
\label{eq:repr}
\end{equation}
where $s_h$ encodes the partial event state, $q_h$ carries
optional verb-prior cues, $u_h$ carries noun-support cues, and
$z_h$ carries global and onset-local features from the frozen
video backbone.

\paragraph{Semantic adapter.}
A lightweight adapter maps $r_h$ to onset and semantic score
parameters:
\begin{equation}
f_\theta(r_h)\rightarrow
(\mu_o,\,\sigma_o^2,\,\ell_o,\,\ell_v,\,\ell_b,\,\ell_n),
\label{eq:adapter}
\end{equation}
where $(\mu_o,\sigma_o^2)$ parameterize the event-local onset
estimate, and $\ell_o,\ell_v,\ell_b,\ell_n$ are raw scores for
onset support, verb, noun existence, and noun, respectively. These
quantities are transformed into event-local support distributions
after normalization and decoding. Onset and semantics are predicted
jointly from the same partial-state representation rather than in
isolated modules.

\paragraph{Statistics-guided cooperative refinement.}
Because onset, verb, noun existence, and noun are mutually
informative, we refine the coarse adapter output with a
deterministic \emph{Statistics-guided Cooperative Refinement (SCR)}
operator rather than a second trainable network. Let $S$ denote the
statistics bundle accumulated from interaction history, including
verb-conditioned and noun-conditioned onset priors, verb--noun
co-occurrence rates, and verb-specific no-noun tendencies.
Refinement is written as
\begin{equation}
(\tilde{\ell}_o,\tilde{\ell}_v,\tilde{\ell}_b,\tilde{\ell}_n)
=\mathcal{R}(\ell_o,\ell_v,\ell_b,\ell_n;\,S),
\label{eq:refine}
\end{equation}
where $\mathcal{R}$ performs deterministic fixed-weight reweighting
and blending of the coarse scores using empirical priors and
co-occurrence statistics, followed by normalization in dependency
order: noun existence, noun, verb, and then onset.

\paragraph{Feature-feedback re-decoding.}
Let $\hat{y}_h^{(1)}$ denote the event hypothesis decoded from the
refined outputs
$(\tilde{\ell}_o,\tilde{\ell}_v,\tilde{\ell}_b,\tilde{\ell}_n)$.
An injection operator $G$ writes selected components of
$\hat{y}_h^{(1)}$ back into the event-state feature slots, and the
adapter is re-applied to obtain a second set of raw scores:
\begin{equation}
r_h'=G(r_h,\hat{y}_h^{(1)}),\qquad
f_\theta(r_h')\rightarrow
(\mu_o',\,{\sigma_o'}^2,\,\ell_o',\,\ell_v',\,\ell_b',\,\ell_n').
\label{eq:feedback}
\end{equation}
A second-pass hypothesis $\hat{y}_h^{(2)}$ is then decoded from
these scores following the same procedure as
Eq.~\eqref{eq:adapter}. The refined pass is accepted only when it
improves the structured joint score, so this step acts as a
consistency correction rather than an unconstrained overwrite.
Because backbone features are frozen and reused, online refresh
requires only a coarse event-local pass followed, when beneficial,
by a small number of lightweight refinement passes.

\subsection{Human-Lock Clamp-and-Re-decode}
\label{sec:clamp}

The system performs \emph{completion} of a partial event, not
relabeling from scratch. Human-confirmed fields must therefore be
preserved exactly under any subsequent automatic refresh, and this
requirement must enter the decoding problem directly rather than as
a soft regularizer.

Let $\mathcal{L}_h$ be the set of locked variables, and let
$\bar{y}_{h,j}$ denote the confirmed value of each
$j\in\mathcal{L}_h$. The feasible set $\mathcal{Y}_h$ clamps all
locked variables to their confirmed values and enforces ontology
validity. Decoding is then defined as
\begin{equation}
\hat{y}_h=\arg\max_{y\in\mathcal{Y}_h}\mathcal{J}_h(y),
\label{eq:decode}
\end{equation}
where $\mathcal{J}_h(y)=J_{\mathrm{onset}}(t_o)+J_{\mathrm{verb}}(v)
+J_{\mathrm{nex}}(b)+J_{\mathrm{noun}}(n)
+J_{\mathrm{compat}}(v,n,t_o)$ sums the event-local log-probability
scores for onset, verb, noun existence, and noun with a structured
onset--verb--noun compatibility term. Any configuration that would
alter a locked field, assign an invalid noun, or realize a noun-free
verb that requires one is excluded by construction through
$\mathcal{Y}_h$.

\subsection{Trust-Calibrated Supervisory Control}
\label{sec:control}

A valid completion does not determine system behavior by itself.
The supervisory question remains: should the system query the
annotator directly, present a suggestion for confirmation, or apply
a conservative local completion autonomously? We address this with
a \emph{Trust-Calibrated Supervisory Controller (TSC)} operating
on a structured control state
\begin{equation}
c_h=\chi(x_h,\hat{y}_h),
\label{eq:ctrl_state}
\end{equation}
which aggregates field completion status, evidence gaps, suggestion
provenance, decoded confidence summaries, and overwrite-risk
context. Each candidate intervention $\xi\in\mathcal{Q}(c_h)$
specifies a target field or field bundle, an interaction surface,
and an authority level.

Candidate-conditioned heuristic estimates of propagation gain
$P(\xi;c_h)$, interaction cost $C(\xi;c_h)$, and overwrite risk
$R(\xi;c_h)$ are adjusted using empirical acceptance and cost
statistics collected from prior interaction history. The resulting
calibrated quantities are denoted by $\bar{P}(\xi;c_h)$,
$\bar{C}(\xi;c_h)$, and $\bar{R}(\xi;c_h)$. Candidates are then
ranked by
\begin{equation}
S(\xi;\,c_h)
=\lambda_1 U(\xi;c_h)+\lambda_2\bar{P}(\xi;c_h)
-\lambda_3\bar{C}(\xi;c_h)-\lambda_4\bar{R}(\xi;c_h),
\label{eq:score_tsc}
\end{equation}
where $U(\xi;c_h)$ measures immediate utility. We enumerate all
candidates in $\mathcal{Q}(c_h)$ and remove any that violate a lock
constraint or exceed the current authority policy, yielding the
executable set $\mathcal{Q}_{\mathrm{safe}}(c_h)$. The selected
intervention is then
\begin{equation}
\xi_h^\star
=\arg\max_{\xi\in\mathcal{Q}_{\mathrm{safe}}(c_h)}S(\xi;\,c_h).
\label{eq:opt}
\end{equation}
A semantically plausible completion may therefore be downgraded
from autonomous execution to human confirmation, or from
confirmation to direct manual query, when the available supervisory
evidence does not justify stronger machine initiative.

\subsection{Human-Machine Supervisory Loop}
\label{sec:loop}

The full annotation process runs as a closed loop. At step $k$,
the controller selects $\xi_h^{(k)}$, presents it in the
appropriate interaction form, and receives annotator response
$a_h^{(k)}$. The event state then evolves as
\begin{equation}
x_h^{(k+1)}
=\mathcal{T}\!\left(x_h^{(k)},\,\xi_h^{(k)},\,a_h^{(k)}\right).
\label{eq:transition}
\end{equation}
Accepted and edited responses promote fields to confirmed status,
thereby narrowing the machine-editable region of the event for all
subsequent decoding passes. Human correction remains possible at
every step, and any autonomous action later overridden by the
annotator is logged with full provenance, providing the empirical
traces required for controller recalibration.
\section{Experiments}
\label{sec:experiments}

We evaluate IMPACT-HOI from four perspectives: annotation efficiency, final event quality, supervisory behavior, and execution safety. As a mixed-initiative framework for structured HOI construction, our evaluation focuses not only on the correctness of final annotations but also on how human effort is allocated and how safely machine assistance operates during event construction.

\subsection{User Study Protocol}
\label{sec:user_study}

We conduct a within-subject user study with 9 participants under two conditions: \texttt{Manual} and \texttt{Full Assist}. In \texttt{Manual}, participants construct HOI events without system assistance. In \texttt{Full Assist}, the full framework is enabled, including HOP-based onset guidance, LPEC-based event completion, SCR-based refinement, and TSC-based intervention selection with lock-preserving execution. Condition order is counterbalanced.

We use a curated subset of procedural HOI clips with reference annotations. To maintain a controlled and reproducible protocol, we restrict evaluation to single-person clips with default \texttt{Left}/\texttt{Right} hand entities, predominantly single-active-hand interactions, and relatively clear temporal and semantic structure. Each participant first completes a short training phase on held-out clips and then annotates 15-second video clips per condition. A trial is complete when all intended HOI events have been entered, the required fields have been resolved, and the annotation can be saved successfully. All sessions are logged to support both user-centered and controller-level analysis, including intervention type, authority level, annotator response, and the empirical traces used for behavioral and safety evaluation.

\subsection{Evaluation Metrics}
\label{sec:exp_metrics}

\noindent\textbf{Efficiency.}
We measure total annotation time per clip and human actions per event. For a clip $c$, total annotation time is
\begin{equation}
T(c)=t_{\text{final save}}(c)-t_{\text{trial start}}(c).
\end{equation}

\noindent\textbf{Final event quality.}
Participant annotations are matched to reference events within each clip by hand identity and temporal overlap. We report onset error, temporal IoU, verb accuracy, noun accuracy, and complete event match rate. Onset error is
\begin{equation}
E_{\text{onset}}=|\hat{t}_o-t_o^*|,
\end{equation}
and temporal IoU is
\begin{equation}
\mathrm{tIoU}=
\frac{|[\hat{t}_s,\hat{t}_e]\cap[t_s^*,t_e^*]|}
{|[\hat{t}_s,\hat{t}_e]\cup[t_s^*,t_e^*]|}.
\end{equation}
A complete event match requires onset correctness within a tolerance band, sufficient temporal overlap, and correct verb and noun labels, with $\delta_o=5$ and $\tau=0.5$.

\noindent\textbf{Supervisory behavior.}
We report suggestion acceptance rate, correction rate, query-type distribution, and authority-level distribution. Acceptance rate is
\begin{equation}
\mathrm{AcceptRate}
=
\frac{N_{\text{accepted suggestions}}}
{N_{\text{accepted suggestions}}+N_{\text{rejected suggestions}}},
\end{equation}
and correction rate is
\begin{equation}
\mathrm{CorrectionRate}
=
\frac{N_{\text{accepted suggestions later edited}}}
{N_{\text{accepted suggestions}}}.
\end{equation}

\noindent\textbf{Safety.}
We report rework after automation, rollback count, and confirmed-field violation count, directly evaluating whether efficiency gains are achieved without unsafe modification of trusted event content.

\subsection{Main User Study Results}
\label{sec:exp_main_results}

Table~\ref{tab:main_results} compares \texttt{Manual} and \texttt{Full Assist}. Our central question is whether supervisory assistance reduces human effort while preserving event quality. We focus on annotation actions per event, complete event match rate, and confirmed-field safety as the key indicators.

\begin{table}[t]
\centering
\caption{Main user study comparison between \texttt{Manual} and \texttt{Full~Assist}.
Metrics are normalized to per-event annotation effort.}
\vskip-1ex
\label{tab:main_results}
\setlength{\tabcolsep}{5pt}
\renewcommand{\arraystretch}{1.15}
\resizebox{\columnwidth}{!}{%
\begin{tabular}{@{}lcccc@{}}
\toprule
\textbf{Method}
  & \textbf{Time (s) $\downarrow$}
  & \textbf{Act./Evt. $\downarrow$}
  & \textbf{Onset Err. (frames) $\downarrow$}
  & \textbf{Complete (\%) $\uparrow$} \\
\midrule
Manual      & 80.13 & 13.33 & 7.06 & 42.22 \\
Full Assist & 90.40 & 11.53 & 9.14 & 46.67 \\
\bottomrule
\end{tabular}%
}
\vskip-1ex
\end{table}

\noindent\textbf{Interaction Burden Reduction.}
\textit{Full Assist} reduces user operations from 13.33 to 11.53 actions per event, confirming that the system shifts annotators from repetitive low-level editing toward high-level decision making: a single \textit{Accept} action can replace several conventional steps, including temporal navigation, label assignment, and field refinement.
Total annotation time increases moderately from 80.13\,s to 90.40\,s, consistent with the changed nature of the task—users are performing substantive proposal-validation steps rather than rapid mechanical operations. This trade-off is the intended outcome of a supervisory workflow.

\noindent\textbf{Annotation Quality.}
Despite fewer manual actions, \textit{Full Assist} improves the Complete Event Match Rate from 42.22\% to 46.67\%, indicating that structured assistance enhances end-to-end annotation quality rather than merely reducing interaction count. While onset error increases slightly (7.06 vs.\ 9.14 frames), the improvement in complete event-level correctness suggests that global consistency benefits outweigh small temporal deviations.

\noindent\textbf{Safety of Supervisory Automation.}
The framework recorded 95 accepted automated operations with zero confirmed-field violations across all logged interactions. This is consistent with the trust-calibrated execution design: by construction, the controller excludes any candidate that would overwrite a confirmed field, and no violation was observed under the studied protocol. For safety-sensitive industrial annotation workflows, this property is essential: automation must remain helpful while preserving the integrity of human-approved decisions.

\begin{figure*}[t]
    \centering
    \subfloat[Efficiency and final event quality.\label{fig:main_results_a}]{%
        \includegraphics[height=3.3cm]{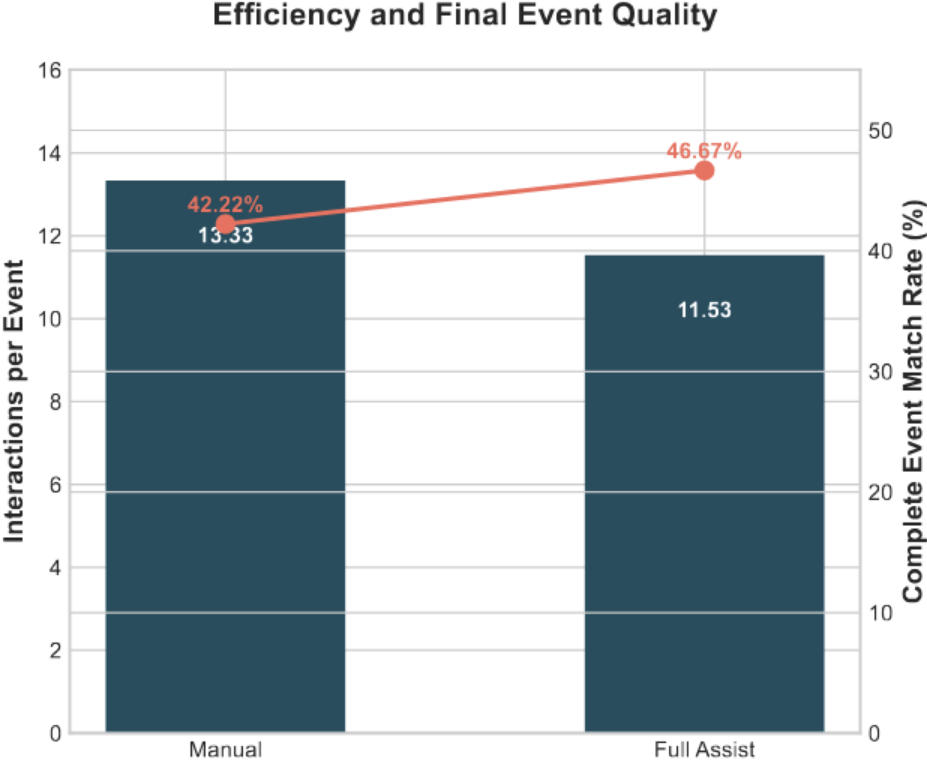}%
    }
    \hspace{0.01\textwidth}
    \subfloat[Supervisory behavior and safety.\label{fig:main_results_b}]{%
        \includegraphics[height=3.3cm]{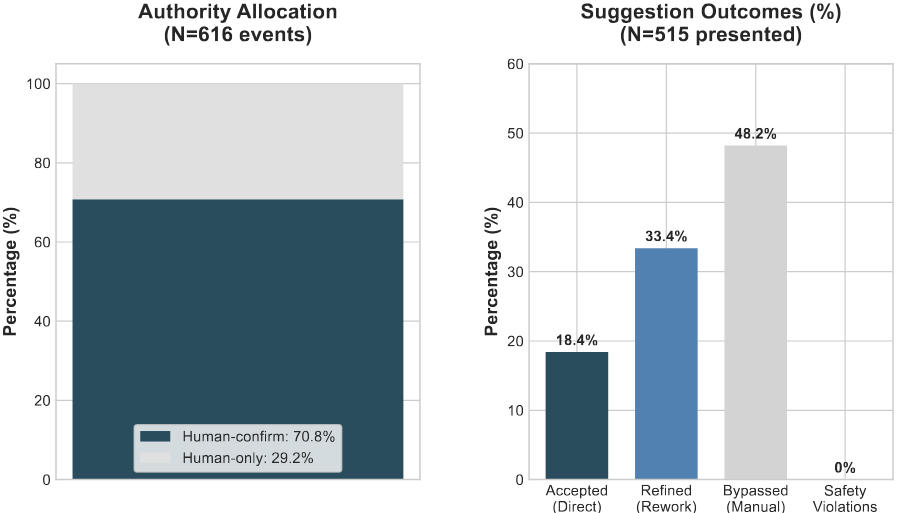}%
    }
    \hspace{0.01\textwidth}
    \subfloat[Comparison of workload.\label{fig:nasatlx}]{%
        \includegraphics[height=3.3cm]{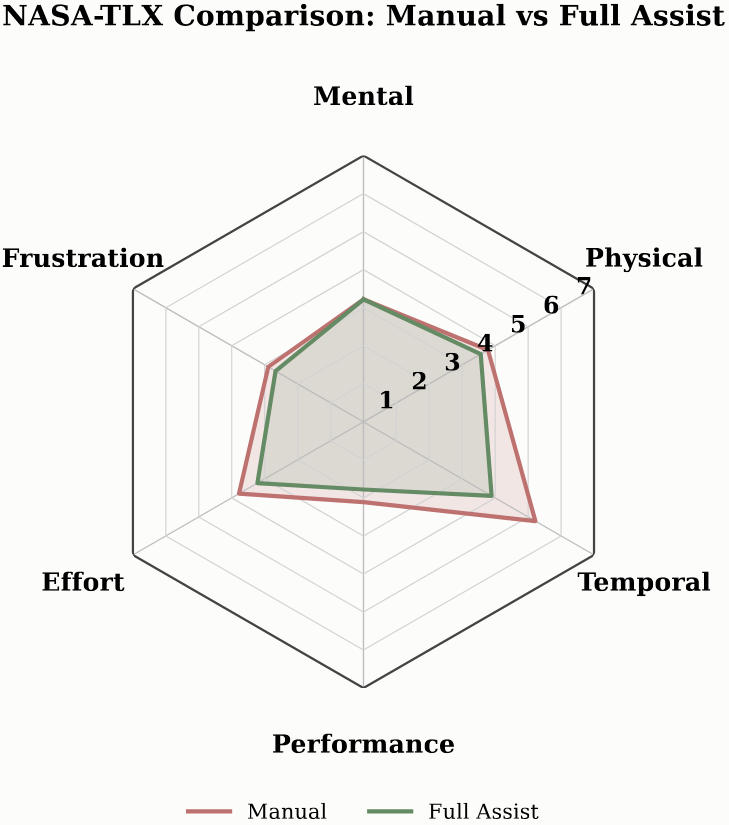}%
    }
    \caption{Main user study results of IMPACT-HOI. (a) shows efficiency and final event quality. (b) shows supervisory behavior and safety outcomes. (c) compares subjective workload between \texttt{Manual} and \texttt{Full~Assist} across the six NASA-TLX dimensions. The radial scale (1--7) indicates the intensity of each demand; a smaller enclosed area suggests a more balanced human--AI collaborative workflow.}
    \label{fig:main_results}
\end{figure*}

Figure~\ref{fig:main_results} visualizes the main outcomes.

\subsection{Supervisory Behavior and Safety}
\label{sec:exp_behavior}

\begin{table}[t]
\centering
\caption{Behavioral analysis of the trust-calibrated supervisory
controller under \texttt{Full Assist}.}
\vskip-2ex
\label{tab:behavior_results}
\setlength{\tabcolsep}{8pt}
\renewcommand{\arraystretch}{1.1}
\begin{tabular}{lcc}
\toprule
Metric & Count & \% \\
\midrule
\multicolumn{3}{l}{\textit{Suggestion Utility ($N_s = 515$ suggestions)}} \\
Accepted suggestions       & 95  & 18.4\% \\
Rework after suggestion    & 172 & 33.4\% \\
Confirmed-field violations & 0   & 0.0\%  \\
\midrule
\multicolumn{3}{l}{\textit{Authority Allocation ($N_o = 616$ operations)}} \\
Human-confirm actions      & 436 & 70.8\% \\
Human-only actions         & 180 & 29.2\% \\
\bottomrule
\end{tabular}
\vskip-1ex
\begin{flushleft}
\footnotesize
Accepted suggestions are directly approved by the annotator; rework after suggestion refers to user refinement of an automated proposal.
\end{flushleft}
\vskip-1ex
\end{table}

To complement the aggregate results, we analyze the internal behavior of the supervisory controller under \textit{Full Assist}. Table~\ref{tab:behavior_results} summarizes suggestion outcomes, authority allocation, and safety-critical events, revealing how responsibility is shared between the human annotator and the automated assistant.

\noindent\textbf{Authority Allocation and Human-Centered Supervision.}
Of the 616 total operations, 436 were executed under \textit{Human-confirm} mode, while 180 remained \textit{Human-only}. This distribution is intentional: the controller provides grounded proposals while the annotator retains final authority through explicit confirmation, preserving accountability while reducing repetitive low-level tasks.

\noindent\textbf{Suggestion Utility and Expert-in-the-Loop Refinement.}
The controller generated 515 suggestions, of which 95 were directly accepted; 172 rework actions followed automated proposals. This behavior reflects a productive expert-in-the-loop workflow in which users frequently treat system outputs as structured initialization and then apply fine-grained adjustments. In complex HOI annotation, partial reuse of a proposal reduces initialization effort while preserving expert precision.

\noindent\textbf{Safety and Conservative Control.}
The framework recorded zero confirmed-field violations throughout the study. The trust-calibrated controller activates autonomous assistance only when confidence and authority constraints are both satisfied, consistent with its role as a conservative risk gate.

\noindent\textbf{Implications for Human--Autonomy Teaming.}
Overall, IMPACT-HOI functions as a collaborative assistant rather than a replacement annotator. The system provides frequent actionable proposals; the human remains the final decision maker; and execution safeguards preserve annotation integrity throughout. This operating model is well-suited for safety-sensitive industrial annotation, where efficiency gains must not compromise data reliability.

\subsection{Ablation Study}
\label{sec:exp_ablation}

To assess IMPACT-HOI's components, we use a \textit{Sequential Oracle-Correction} protocol: the system predicts HOI onset, verb, and noun in sequence; if a prediction deviates from ground truth beyond a tolerance ($\delta_o=5$ frames for onset, exact match for semantics), an oracle corrects the field, locks it, and the system re-decodes the remaining fields. This measures the framework's effectiveness in reducing human correction effort while respecting prior decisions.
We use a VideoMAE (ViT-B) encoder pretrained on Kinetics-710 for feature extraction. The semantic adapter is trained for \textbf{1 epoch} on 26 clips (${\sim}35{,}000$ event samples). Evaluation is conducted on 9 held-out clips, with supervisory statistics estimated from the training set.
Table~\ref{tab:ablation_results} reports both predictive and operational metrics: \textit{Action Reduction} (decrease in required manual actions) and \textit{Zero-Edit Rate} (events accepted without human modification). These metrics reflect practical value more directly than recognition accuracy alone.

\begin{table}[t]
\centering
\caption{Ablation Study of IMPACT-HOI components, evaluated across 1,235 events
under Sequential Oracle-Correction.
O.~MAE: Onset Mean Absolute Error;
O.~Acc@5: Onset Accuracy within 5 frames;
Verb/Noun: Top-1 accuracy after GT onset grounding;
Action Red.: reduction in human actions;
Zero-Edit: rate of events needing no correction.
TSC modulates interaction authority, not offline predictive accuracy.}
\label{tab:ablation_results}
\resizebox{\columnwidth}{!}{%
\begin{tabular}{lcccccc}
\toprule
\textbf{Variant}
  & \textbf{O.~MAE $\downarrow$}
  & \textbf{O.~Acc@5 $\uparrow$}
  & \textbf{Verb $\uparrow$}
  & \textbf{Noun $\uparrow$}
  & \textbf{Action Red. $\uparrow$}
  & \textbf{Zero-Edit $\uparrow$} \\
\midrule
\textit{Baselines} \\
Semantic Only              & 12.94 & 50.6\% & 39.5\% &  9.6\% & 33.3\% &  1.8\% \\
\textbf{Full Assist (Ours)}& 10.88 & \textbf{55.8\%} & \textbf{59.8\%} & 86.2\% & 67.2\% & \textbf{33.3\%} \\
\midrule
\textit{Ablation Variants} \\
w/o HOP (No Motion Prior)  & \textbf{10.77} & 55.5\% & \textbf{59.8\%} & 86.2\% & 67.2\% & 33.2\% \\
w/o SCR (No Stat Refine)   & 11.82 & 51.7\% & 59.1\% & 86.6\% & 65.8\% & 30.5\% \\
w/o Lock-Aware Decoding    & 10.88 & \textbf{55.8\%} & 58.0\% & \textbf{88.9\%} & \textbf{67.6\%} & 32.9\% \\
\bottomrule
\end{tabular}%
}
\vskip-3ex
\end{table}

\paragraph{Establishing System Necessity: Baseline vs.\ Full Assist.}
The appearance-based baseline (\textit{Semantic Only}), relying solely on global VideoMAE features, achieves only 9.6\% noun accuracy, rendering autonomous proposals unsuitable for the fine-grained tool-and-component discrimination required in industrial assembly. The resulting Zero-Edit Rate of 1.8\% confirms that nearly every proposal requires manual intervention. In contrast, \textit{Full Assist} raises noun accuracy to 86.2\% and Zero-Edit Rate to 33.3\%, demonstrating that the mixed-initiative framework effectively shifts the human role from data entry to high-level semantic supervision.

\paragraph{Temporal Anchoring and the HOP Prior.}
Ablating HOP (\textit{w/o HOP}) marginally reduces average onset error (10.88\,$\to$\,10.77 frames) but lowers O.~Acc@5 from 55.8\% to 55.5\%. In mixed-initiative annotation, O.~Acc@5 is the more critical measure: average error can be skewed by long-tail outliers, such as occlusion-induced drift, whereas O.~Acc@5 directly reflects whether the system lands within the precision band that avoids manual correction. HOP's hand-motion evidence anchors predictions to functional contact points, sustaining this hit rate and thereby preserving Zero-Edit instances.

\paragraph{Logical Consistency through SCR and Lock-Aware Decoding.}
Ablating Lock-Aware Decoding (\textit{w/o Lock-Aware}) yields slightly higher noun accuracy (88.9\% vs.\ 86.2\%), but this is a deceptive gain: without conditioning on human-confirmed verb labels, the unconstrained model optimizes $P(\text{Noun})$ rather than $P(\text{Noun}\mid\text{Verb}_{\text{GT}})$, occasionally recovering correct nouns by ignoring explicit user decisions. The resulting drop in Zero-Edit Rate (33.3\%\,$\to$\,32.9\%) confirms the interaction cost of this inconsistency.
Ablating SCR (\textit{w/o SCR}) worsens Onset MAE from 10.88 to 11.82 frames and reduces Zero-Edit Rate to 30.5\%, demonstrating its role in coupling semantics and temporality. Together, SCR and Lock-Aware Decoding stabilize the system against representational uncertainty, ensuring that efficiency gains arise from reliable structured assistance rather than unconstrained prediction.

\paragraph{Trust Calibration as an Interaction Risk Gate.}
TSC does not affect offline predictive accuracy under oracle-correction—by design, oracle correction eliminates interaction-level errors, so Table~\ref{tab:ablation_results} does not include a \textit{w/o TSC} row for predictive metrics. TSC's contribution is supervisory risk control. Without TSC, the uncalibrated system attempts over 700 autonomous semantic applications in the one-shot phase, of which approximately 50 are incorrect. In real annotation, this overconfident autonomy forces users to hunt and revert unsolicited errors, degrading trust. With TSC, these high-risk interventions are intercepted and downgraded from autonomous execution to human-confirmation prompts, confining autonomy to high-confidence regimes without sacrificing coverage.

\paragraph{Synthesis: Achieving Mixed-Initiative Autonomy.}
\textit{Full Assist} achieves a 67.2\% Action Reduction, automating over two-thirds of low-level labeling steps relative to a manual workflow. The 33.3\% Zero-Edit Rate indicates that one-third of all events require no human modification within the defined tolerances. These gains arise from the synergistic integration of kinematic priors (HOP), statistical refinement (SCR), and structured conditional reasoning (Lock-Aware Decoding), with TSC ensuring that the system's initiative remains within authority-appropriate boundaries.

\subsection{Statistical Analysis and Qualitative Results}
\label{sec:exp_stats_qual}

To quantify the impact of IMPACT-HOI on subjective workload, we conducted a within-subject analysis based on the NASA-TLX scale. As visualized in Fig.~\ref{fig:nasatlx}, \textit{Full Assist} showed lower scores than \texttt{Manual} across five of six dimensions.

\noindent\textbf{Significant Reduction in Temporal Demand.}
The most prominent finding is a statistically significant reduction in \textit{Temporal Demand} ($p = 0.0157$, Wilcoxon signed-rank; Manual mean\,=\,5.22, Full Assist mean\,=\,3.89). In manual mode, participants reported sustained pressure from continuous frame-seeking and category-matching. Automated onset completion and structured suggestions in IMPACT-HOI mitigated this pressure, allowing users to focus on quality auditing rather than rhythmic speed-matching.

\noindent\textbf{Cognitive Stability.}
Despite the supervisory loop, \textit{Mental Demand} remained stable (mean\,=\,3.22 for both conditions), indicating that the trust-calibrated interface delivers assistance without increasing information-processing overhead. Positive trends are observed in \textit{Performance} (lower failure perception: 1.78 vs.\ 2.11) and \textit{Effort} (3.22 vs.\ 3.78), though these do not reach significance in our sample.

\noindent\textbf{Qualitative Feedback.}
Most participants noted that manual annotation felt like a ``low-level coordinate hunt,'' while the collaborative mode felt like a ``knowledge-level verification,'' consistent with the design goal of shifting the human role from recorder to supervisory expert.
\section{Conclusion}
In this paper, we introduced IMPACT-HOI, a mixed-initiative
framework for efficient and reliable Human-Object Interaction
(HOI) annotation. By incrementally resolving a partially
specified state, IMPACT-HOI improves complete event match
rate and reduces manual annotation actions. The
Trust-Calibrated Supervisory Controller (TSC) bounds machine
initiative to high-confidence, authority-compliant regimes,
protecting human-confirmed decisions throughout. Our user
study shows a 13.5\% reduction in annotation actions and a
46.67\% event match rate with zero confirmed-field violations
under the studied protocol. These results support IMPACT-HOI
as an annotation infrastructure for structured, safety-sensitive
workflows of the kind required for learning robot manipulation
from human demonstration.

\bibliographystyle{unsrt}
\bibliography{bib.bib}

\end{document}